\documentclass[sigconf]{acmart}

\usepackage{booktabs} 
\usepackage{forest}
\usepackage{multicol}
\usepackage{bbm}

\tikzset{
    basic/.style  = {draw, text width=2cm, drop shadow, rectangle},
    root/.style   = {basic, rounded corners=2pt, thin, align=center, fill=green!30},
    level 2/.style = {basic, rounded corners=6pt, thin,align=center, fill=green!60, text width=8em},
    level 3/.style = {basic, thin, align=left, fill=pink!60, text width=6.5em},
    every label/.append style={font=\scriptsize},
    my edge labels/.style={font=\scriptsize},
    dominant/.append style={label=below:$dominant$},
}

\setcopyright{acmlicensed}

\acmDOI{xx.xxx/xxx_x}

\acmISBN{978-1-4503-6866-7/20/03}

\acmConference[]{}{}{} 
\acmYear{2020}
\copyrightyear{2020}

\acmArticle{4}
\acmPrice{15.00}

\begin{document}
\title{Global Hierarchical Neural Networks using Hierarchical Softmax}



\author{Jetze Schuurmans}
\affiliation{%
  \institution{Erasmus University Rotterdam}
  \streetaddress{Burgemeester van Walsumweg 50}
  \city{Rotterdam} 
  \country{Netherlands}
  \postcode{3063}
}
\email{jetzeschuurmans@gmail.com}

\author{Flavius Fr\u{a}sincar}
\affiliation{%
  \institution{Erasmus University Rotterdam}
  \streetaddress{Burgemeester van Walsumweg 50}
  \city{Rotterdam} 
  \country{Netherlands}
  \postcode{3063}
}
\email{frasincar@ese.eur.nl}

\renewcommand{\shortauthors}{J. Schuurmans et al.}

\begin{abstract}
This paper presents a framework in which hierarchical softmax is used to create a global hierarchical classifier. The approach is applicable for any classification task where there is a natural hierarchy among classes. We show empirical results on four text classification datasets. In all datasets the hierarchical softmax improved on the regular softmax used in a flat classifier in terms of macro-F1 and macro-recall. In three out of four datasets hierarchical softmax achieved a higher micro-accuracy and macro-precision. 
\end{abstract}

%
%

\ccsdesc[500]{Computing methodologies~Supervised learning by classification}
\ccsdesc[500]{Computing methodologies~Neural networks}

\keywords{Hierarchical Classification, Hierachical Softmax, Global Hierarchy}

\maketitle

\section{Introduction}
In machine learning, classification is a popular and well-studied problem. 
Recently neural networks have flourished in different classification tasks, especially when the number of training examples is large. 
In some cases the classes have a natural taxonomy, creating a hierarchy among them. Hierarchical classification tries to incorporate this hierarchy in the classifier, as apposed to flat classifiers, which need to discriminate between all classes at once. 

There are two different types of hierarchical classifiers, local and global. The local classifier is a combination of flat classifiers, while the global classifier adapts the classifier internally. When large neural networks are trained it is often unrealistic to train multiple networks, due to computation time, and recreate the hierarchy of the classes by using multiple local classifier. Therefore we use a global classifier. We do this by exchanging the softmax with a hierarchical softmax, such that any Neural Network can be modified to a hierarchical classifier. We show that this adjustment makes the network a truly global hierarchical classifier and that it can enhance the performance in several classification tasks.\footnote{The code for reproducing the results can be found at: \url{https://github.com/JSchuurmans/hSoftmax}.}

The paper is structured as follows. In Section \ref{sec:prev} previous works on hierarchical classifiers and hierarchical softmax is covered. Our proposal for the hierarchical softmax is presented in Section \ref{sec:meth}. Then in Section \ref{sec:data} we describe several datasets and Section \ref{sec:exp} discusses the experimental setup. In Section \ref{sec:res} we compare the results of models with a regular softmax and with a hierarchical softmax on these datasets. Finally, in Section \ref{sec:conc} we give our conclusions and proposals for future work.

\section{Previous Work on Hierarchical Classification} \label{sec:prev}
Most classification algorithms could be considered \textit{flat classifiers}. They distinguish between all classes at once. When there are a large number of classes, this can become difficult. Instead, \textit{hierarchical classification} can be used. A hierarchical classifier tries to incorporate the hierarchical structure of the class taxonomy, when this is present. 

\cite{silla2011survey} proposed a survey on hierarchical classification and built a unifying framework for distinguishing methods. First the structure of the taxonomy is considered, which can be a Tree or a Directed Acyclical Graph (DAG). Working with trees is easier, because a node in a DAG can have more than one parent node. With regard to the classifier, this can be local (top-down) or global (big-bang). 

The local hierarchical classifier is not a full hierarchical method on its own. Instead it is a group of flat classifiers that during training considers a subset of classes, based on where in the taxonomy the flat classifier is used \citep{silla2011survey}. A global classifier takes the hierarchy of the classes into a single model \citep{freitas2007tutorial}. The advantages of a global classifier are a smaller model and class dependencies are automatically incorporated \citep{silla2011survey}. The global classifier also has the convenience that the number of parameters is far less than for the same classifier in a local hierarchy. More importantly, a misclassification at a certain level is unrecoverable in a local hierarchy, while in a global hierarchy this can be compensated.


\subsection{Global Hierarchy}

There are different types of global classifiers. First, there are global approaches based on the approach of \cite{rocchio1971relevance} that use class clusters. 
An example specific to text mining is given in \citep{labrou1999yahoo}. The second type of global classifiers are built on the multi-label classification. In this approach the non-leaf nodes are supplemented with the information of their parent nodes. 
Finally, the last type of a global classifier is a modification of a local classifier to incorporate the class hierarchy directly. Although harder to construct, the output of this last method might be easier to explain than the one from a local based approach \cite{silla2011survey}. During both training and testing probabilities of all classes can be assessed. 

\subsection{Hierarchical Softmax}
Hierarchical softmax was first described by \cite{goodman2001classes}. In the context of neural network language models, hierarchical softmax was first introduced in \cite{morin2005hierarchical}. Other versions of hierarchical softmax are proposed in \cite{mikolov2011strategies} and \cite{mnih2009scalable}. \cite{grave2017bag} used the specification from \cite{morin2005hierarchical} in their FastText classifier. 
While most of these methods use a binary tree to speed up training and inference time, we try to exploit the natural hierarchy found in the taxonomy of classes for improving the performance. In this taxonomy a node can have more than two child nodes. 

\subsection{Hierarchical Text Classification}
Hierarchical classification was first used for text classification by \cite{koller1997hierarchically}. They used a local classifier per parent node for training, at each node selecting a subset of features relevant for that step in the classification process. 
A similar hierarchical structure with an SVM at every node was used by \citet{kang2013hierarchical} for speech-act classification. 
\citet{ono2016toward} used a form of local classifier per level, where they tried the lowest level (leaf nodes) first. If the uncertainty was too high, they moved up in the hierarchical level.

\section{Methodology} \label{sec:meth}
Hierarchical classification can be considered as a classification that takes the hierarchical structure of the taxonomy of classes into account, as opposed to a flat classifier, which only takes the final classes into account. By imposing the hierarchical structure, the model does not need to learn the separation between a large number of classes. It can now focus on classifying categories, or subclasses within a category. The taxonomy can be formalised as a tree or a DAG. We consider here the case where the taxonomy is a tree. A taxonomy represented by a Tree is easier to construct, as each child node only has one parent. 

A local hierarchical neural network would be infeasible. The network has many parameters and having to learn a new neural network from scratch at every parent node would result in too many parameters, and considerably longer training and inference times. Therefore, we consider making a global classifier by using a hierarchical softmax. Hierarchical softmax easily extends the neural networks by replacing the regular softmax. 
In this section we discuss the general case of the global hierarchical classifier and the specific case for the hierarchical softmax.


\subsection{Global Hierarchy}
\begin{figure*}[htbp]
    \centering
    \begin{forest}
    for tree={
      circle,
      draw,
      minimum width=2.5em,
      l sep+=1.5em,
      s sep+=1em,
      anchor=center,
      edge path={
        \noexpand\path[\forestoption{edge}](!u.parent anchor)--(.child anchor)[my edge labels]\forestoption{edge label};
      },
    },
  [R, label={left:$P(R)=1$}
    [1, label={left:$P(1)=P(1|R)P(R)=P(1|R)$}, edge label={node[midway,fill=white,font=\scriptsize]{$P(1|R)$}}
      [1.1,  edge label={node[midway,fill=white,font=\scriptsize]{$P(1.1|1)$}},
        label={left:$P(1.1)=P(1.1|1)P(1)=P(1.1|1)P(1|R)$}, label={below:$c_1$}]
      [1.2,  label={below:$c_2$}]
      [\dots,  label={below:\dots}]
    ]
    [2, s sep+=1.5em,
      [2.1, label={below:$c_m$}]
      [\dots, label={below:\dots}]
    ]
    [\dots, s sep+=3em,
        [\dots, label={below:\dots}]
    ]
    [P, s sep+=1.5em,
        [P.1, label={below:\dots}]
        [\dots, label={below:\dots}]
        [P.$J_p$, label={below:C}]
    ]
  ]
  \end{forest}
    \caption{Hierarchical structure of a two level global classifier.}
    \label{fig:global_hier}
\end{figure*}
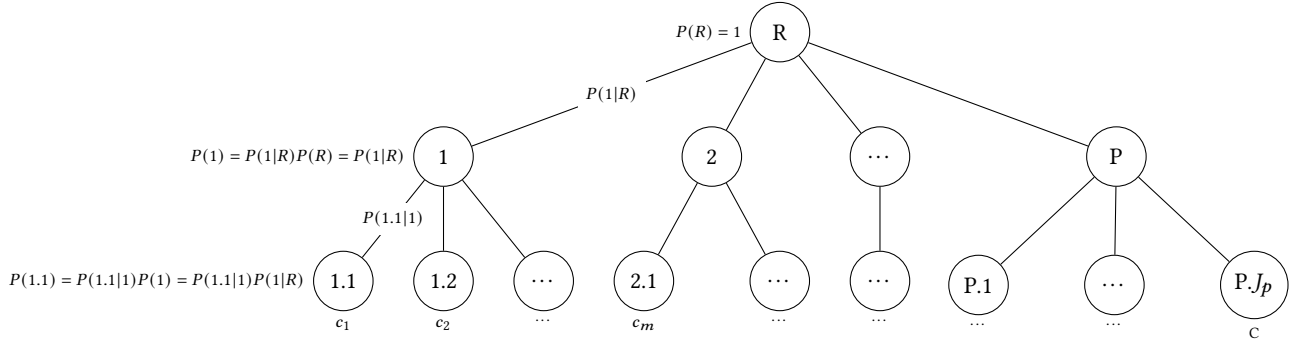

Global classifiers take advantage of the whole hierarchical structure in the classes at once \cite{silla2011survey}. Each node in this hierarchical structure is associated with the probability of the path from the root to that node. We illustrate this in Figure \ref{fig:global_hier}.
If the node is at depth $l$ with parents $n_1, \dots , n_{l-1}$, the probability of arriving in this node is:
\begin{align}
    P(n_{l}) &= P(n_{l}|n_{l-1}) P(n_{l-1}) \\
    &= \prod_{j=1}^{l} P(n_{j}|n_{j-1}). \label{eq:hscd}
\end{align}

\subsection{Hierarchical Softmax}
In order to calculate the conditional probabilities the hierarchical softmax uses a softmax at every node. 
The softmax used in the hierarchical softmax to calculate the conditional probability of belonging to node $m = n_l$ conditional on being in the parent node of $p = n_{l-1}$ becomes:
\begin{equation}
    P(m | p) = \frac{\exp(w_{pm}^T  h)}{\sum_{j=1}^{J_p} \exp(w_{pj}^T  h)},
\end{equation}
where $w_{pm}^T$ is the weight vector corresponding to parent node $p$ and child node $m$. The weight vectors $w$ include the bias terms. $h$ is therefore the last hidden state concatenated with a one, and provides the same input for each parent node, independent of the depth $l$. The number of weight vectors $J_p$ is equal to the number of child classes of parent node $p$. 

Compared to a flat classifier with a regular softmax ($P=1$), the total number of weights does increase with $(P-1)* (h_{dim}+1)$, where $P$ is the number of parent nodes, $h_{dim}$ is the dimension of the hidden dimension and one is added to account for all the additional bias terms. Although the total number of weights is increasing, this is considerably less then if we would consider a new neural network at every parent node, as it is done in the local classifier per parent node. 

Each weight vector now has a new purpose. In a flat classifier with the regular softmax, $\exp(h^T w_j)$ attributes the evidence of class $j$ compared to all leaf nodes $\sum_{c=1}^C \exp(w_c^T h)$, where $C$ equals the number of classes, or leaf nodes. While in the hierarchical softmax, the importance of node $m$, $\exp(w_{pm}^T  h)$, is compared to a subset of nodes $j=1,\dots,J_p$, $\sum_{j=1}^{J_p} \exp(w_{pj}^T  h)$. This gives the hierarchical softmax the potential advantage of only having to make the distinction between smaller subsets. In other words, the additional $(P-1)*(h_{dim}+1)$ parameters empower the $C*(h_{dim}+1)$ parameters to specialise in discriminating within their respective subgroups. 

\subsubsection{Training of the Hierarchical Softmax}
In order to understand the training of a network with a hierarchical softmax component we need to calculate the gradients of the loss function with respect to the parameters, $w_{pj}$ and $h$. This will also show that the hierarchical softmax is truly a global classifier, as the whole network is updated based on the performance of all relevant parent nodes. 

The loss function we use is the Cross Entropy function. For observation $i$ the loss is calculated as a function of the estimated class probabilities $P_i(c)$:
\begin{align}
    E_i &= - \sum_{c=1}^C y_{i,c} \log P_i(c)  \\
    &= - \log P_i(m)  \\
    &= - \log \prod_{q \in Q} P(m_q|q) \label{eq:subst} \\
    &= - \sum_{q \in Q} \log  P(m_q|q) \label{eq:log}
\end{align}
The indicator $y_{i,c}$ is 1 if observation $i$ belongs to class $c$, therefore the element of the sum that remains is the negative log probability of the correct class $m$. We then substitute (\ref{eq:hscd}) in the loss (\ref{eq:subst}), where $Q$ is the set of all parent nodes that lead to the correct class, and $m_q$ the correct child of parent $q$. In (\ref{eq:log}) we rearrange the log, making it easier to calculate the derivatives we are looking for:
\begin{align}
    \label{eq:dE-dw}
    \frac{\partial E_i}{\partial w_{pj}^T } 
    &= \mathbbm{1}_{p \in Q} (P(j|p)-\delta_{jm_p})  h \\
    \frac{\partial E_i}{\partial h } 
    &= \sum_{q \in Q} \sum_{j=1}^{J_q} ( P(j|q) - \delta_{jm_q} ) w_{jq}^T \label{eq:dE-dh}
\end{align}

\noindent where the indicator function $\mathbbm{1}_{p \in Q}$ equals one if $p$ is in $Q$ and zero otherwise. Likewise, the Kronecker delta is defined as $\delta_{jm} = 1$ if $j=m$, and 0 otherwise. The derivations of (\ref{eq:dE-dw}) and (\ref{eq:dE-dh}) are given in Appendix \ref{sec:part}.

These gradients can be used in the Stochastic Gradient Decent algorithm. More importantly, (\ref{eq:dE-dh}) shows the update of the hidden state (and therefore the rest of the network) is a combination of the performances across all child nodes that belong to the parent nodes that make up the path to the correct class. This shows that a neural network with a hierarchical softmax is truly a global hierarchical classifier.

\section{Datasets} \label{sec:data}
We consider four text classification datasets, in which we can find a hierarchical structure in the classes. In Appendix \ref{sec:tax} we present the exact taxonomy used.

\subsection{TREC}
First, we consider the TREC 10 Question Answering Track Corpus \cite{li2002learning}, abbreviated as TREC. This dataset consists of 5952 questions (5452 train, 500 test), each belonging to one of the 50 classes. These classes are split up between 6 categories. Figures \ref{fig:trec} shows that the distribution between categories is highly unbalanced. The TREC training set is highly unbalanced as well. The number of training observations per class range from 4 till 962.

\subsection{20NewsGroups}
The second dataset is the 20NewsGroups dataset \cite{lang1995newsweeder}. We find 6 categories, on top of the 20 classes. In Figure \ref{fig:20ng} the distribution of the classes between the categories can be seen. This distribution is relatively balanced. This dataset contains 11293 training observations and 7527 test observations. The training set is relatively balanced. Most classes have between 500 and 600 observations. There is one outlying class with only 377 training observations.

\subsection{Reuters-21578}
As third and fourth dataset we use two configurations of the Reuters-21578 dataset \cite{hayes1990tcs}. These are Reuters-8 and Reuters-52.

\subsubsection{Reuters-8}
The Reuters-8 dataset consists of the 8 most frequent classes, based on the number of observations in the training set. These are distributed between 4 categories. The classes are distributed evenly between the categories, as shown in Figure \ref{fig:r8}. The observations of this dataset are split between train and test set with 5485 in train and 2189 in test. The training set is very unbalanced, with observations per class ranging from 41 to 2840.

\subsubsection{Reuters-52}
Respectively the Reuters-52 dataset contains the 52 most frequent classes, also distributed between 4 categories. Figure \ref{fig:r52} shows the distribution of classes among the categories is highly unbalanced. The Reuters-52 dataset contains 6532 training and 2568 test observations. This dataset is by definition more unbalanced than Reuters-8. The minimum number is only a single training observation.

\section{Experiments} \label{sec:exp}
In our experiments we employ an LSTM \cite{hochreiter1997long} with hierarchical softmax and compare the results with an LSTM with regular softmax. LSTMs are a popular and well performing architecture for text classification, because of their ability to process sequential data \cite{goldberg2016primer}. Since the meaning of a word might also depend on the words that follow, we also consider the Bidirectional LSTM (BiLSTM) \cite{graves2005framewise}.

\subsection{Hyperparameters}
In order to determine the best hyperparameters we use k-fold cross-validation on the training set, where $k=4$. The macro-F1 measure is used as validation criteria. Besides the Bidirectional component, different dimensions for the hidden state ($h_{dim}$), are tried. The 300 dimensional GloVe \cite{pennington2014glove} word embeddings pretrained on the Wikipedia 2014 and Gigaword 5 (6B tokens) corpus are used. Furthermore, we use early stopping (with a stopping criteria based on cross validation) and dropout (50\%) \cite{zaremba2014recurrent} to prevent overfitting. We train using the Adam optimiser \cite{kingma2014adam} with a learning rate of 0.001, and batch size of 10.

\subsection{Evaluation metrics}
We evaluate the performances based on four metrics, F1, precision, recall, and accuracy. Our main criteria is the F1 measure. The F1 is the harmonic mean of the precision and recall. We value both and do not want a linear trade-off between them. Since we are dealing with unbalanced multi-class classification, the macro-F1 is used. We report macro-precision and macro-recall to give some insight in whether the precision and recall differ and which might be higher or lower. Note that the macro-F1 is in general not the harmonic mean of the macro-precision and macro-recall. Rather, the macro-F1 is the average over the harmonic mean of precision and recall of the individual class. The micro-accuracy is a popular measure used for these datasets \cite{madabushi2016high, wu2019simplifying, yamada2019neural, haonan2019graph}. We include it such that the performance of our models can be compared with other papers.

\section{Results} \label{sec:res}
In all datasets, the hierarchical softmax outperforms the regular softmax in terms of our main criteria, macro-F1. The results are tabulated in Tables \ref{tab:res_trec}-\ref{tab:res_r52}. In this section we discuss the results in more detail.

We find that in all datasets the highest macro-F1 validation scores for a Bidirectional LSTM. The optimal dimension of the hidden state for the regular and hierarchical softmax is the same in all datasets, in most $h_{dim}=150$, while in R-52 the optimal is $h_{dim}=100$.

\begin{table}[htbp] 
    \centering 
    \caption{Performance of BiLSTM $(h_{dim}=150)$ on the TREC dataset.}
    \begin{tabular}{lrrr} 
        \toprule
        Measure & Flat  & Hierarchical & \textit{SOTA}\\ 
        \midrule 
        F1 & 72.595 & \textbf{74.258} & \\
        Precision & 74.937 & \textbf{75.882} & \\
        Recall & 73.370 & \textbf{74.952} & \\
        Accuracy & 84.800 & \textbf{86.000} & \textit{97.2}\\
        \bottomrule
    \end{tabular}
    \label{tab:res_trec}
\end{table}

In the TREC dataset the hierarchical softmax performs better than the regular softmax in terms of all performance measures. The precision is slightly better than the recall for both the hierarchical and flat model. The accuracy is higher than the macro measures, due to the unbalanced classes. Furthermore, the accuracy shows that the hierarchical softmax did not perform as well as the state-of-the-art (SOTA) \cite{madabushi2016high}. 

\begin{table}[htbp] 
    \centering
    \caption{Performance of BiLSTM $(h_{dim}=150)$ on the 20NewsGroups dataset.}
    \begin{tabular}{lrrr}
        \toprule
        Measure & Flat & Hierarchical & \textit{SOTA}\\ 
        \midrule 
        F1 & 81.048 & \textbf{82.479} &\\
        Precision & 81.586 & \textbf{82.979} &\\
        Recall & 80.905 & \textbf{82.465}& \\
        Accuracy & 81.382 & \textbf{82.924} & \textit{88.5}\\
        \bottomrule
    \end{tabular}
    \label{tab:res_20ng}
\end{table}

The hierarchical model also outperforms the flat model in the second dataset, the 20NewsGroups. Here all measures are relatively close, indicating good trade-off between false positives and false negatives. Since the classes are relatively more balanced, the micro-accuracy is for both models closer to the macro measures. The hierarchical model was closer to the state-of-the-art \cite{wu2019simplifying} than the flat model. However, it did not perform as well.

\begin{table}[htbp] 
    \centering
    \caption{Performance of BiLSTM $(h_{dim}=150)$ on the Reuters-8 dataset.}
    \begin{tabular}{lrrr}
        \toprule
        Measure & Flat  & Hierarchical & \textit{SOTA} \\ 
        \midrule 
        F1 & 89.264 & \textbf{92.939} & \\
        Precision & 90.696 & \textbf{93.065} & \\
        Recall & 88.380 & \textbf{93.056} & \\
        Accuracy & 96.711 & \textbf{97.762} & \textit{97.8} \\
        \bottomrule
    \end{tabular}
    \label{tab:res_r8}
\end{table}

In the Reuters-8 dataset the hierarchical softmax outperforms the regular softmax in all performance measures. All measures are relatively close, indicating a good trade-off between false positives and false negatives. Despite the large class unbalance the micro-accuracy of the hierarchical model is relatively close to the macro measures. Furthermore, compared to the state-of-the-art \cite{yamada2019neural}, the hierarchical model is very close to the state-of-the-art.

\begin{table}[htbp] 
    \centering
    \caption{Performance of BiLSTM $(h_{dim}=100)$ on the Reuters-52 dataset.}
    \begin{tabular}{lrrr}
        \toprule
        Measure & Flat  & Hierarchical & \textit{SOTA}\\ 
        \midrule 
        F1 & 63.637 & \textbf{64.407} &\\
        Precision & \textbf{70.325} & 68.361 &\\
        Recall & 62.9258 & \textbf{64.124} &\\
        Accuracy & \textbf{93.575} & 93.380 & \textit{95.00}\\
        \bottomrule
    \end{tabular}
    \label{tab:res_r52}
\end{table}

In terms of our main performance criteria, macro-F1, the hierarchical softmax outperforms the regular softmax in the Reuters-52 dataset. While the recall seems to be the bottleneck, and the hierarchical softmax performs better on the macro-recall, the macro-Precision is higher for the regular softmax. The larger difference between recall and precision indicate a worse trade-off in the flat model. With a difference of 0.2 percentage points, the micro-accuracy of the regular softmax is not much higher than the micro-accuracy of the hierarchical softmax. The flat and hierarchical models have a larger gap between the micro-average and macro measures, due to the higher class unbalance. Finally, we note that both and are close to the state-of-the-art \cite{haonan2019graph}.

\section{Conclusion} \label{sec:conc}
We conclude that the hierarchical softmax makes a good candidate for making a neural network a global hierarchical classifier. We show that it can improve performances of a recurrent network on four different text classification datasets, in terms of macro-F1 and macro-Recall. The performances on the different datasets show that the hierarchical softmax can handle different types of class taxonomies, balanced and unbalanced, in terms of both training observations per class, as well as child nodes per parent node. 

With regard to the state-of-the-art, it is not our goal to improve the SOTA, instead we show that changing a regular softmax with a hierarchical softmax in a dataset with a natural hierarchy in the classes leads to an improvement. Although we did not improve the state-of-the-art, we do come close with the hierarchical softmax on a parsimonious model. We also note that the SOTA models are a different model for each dataset, while we consistently perform well on all datasets with the same model. Future work can study if the state-of-the-art might be improved if the hierarchical softmax is used in the respective model.

Furthermore, we consider a two-level hierarchical taxonomy, by introducing one level of parent nodes in between the root and the leaves. In future work, the taxonomy could be extended with an additional hierarchical layer, i.e. by grouping parent nodes. 

The hierarchy is currently determined based on the hierarchy in the class taxonomy. Alternatively the construction and evaluation of different hierarchical structures could be automated.

The performance of the hierarchical softmax depends on the probability estimates of the conditional probabilities of moving from a parent node to a child node. Better estimates might be obtained by using Bayesian neural networks, as their probability estimates are significantly better \cite{fortunato2017bayesian, lipton2018bbq, siddhant2018deep}.

The hierarchical softmax is not only applicable for text classification. In theory it can replace a softmax in any classification task. It would also be interesting to see how this approach fares in other classification tasks, for example in image classification.

\bibliographystyle{ACM-Reference-Format}
\bibliography{bibliography} 

\appendix

\section{Derivation of the derivatives} \label{sec:part}
In this Appendix we derive the derivatives (\ref{eq:dE-dw}) and (\ref{eq:dE-dh}). We start with the derivative with respect to the weights, then follows the derivative with respect to the hidden state.

In both derivations we use the derivative of the softmax estimate of the probability of node $k$, $P(k|p)$, with respect to the inner product of the weight vector and the hidden state $w_{pj}^T h$:

\begin{align}
    \frac{\partial P(k|p)}{\partial w_{pj}^T h} &=  P(k|p) ( \delta_{jk} -P(j|p))
\end{align}

\subsection{Derivative with respect to the weights}
We calculate the derivative of the Cross Entropy loss of observation $i$ with respect to a given weight vector $w_{pj}^T$ to show that the weight updates are relatively straight forward. The derivative can be split up using the chain rule:

\begin{align}\label{eq:chain_w}
    \frac{\partial E_i}{\partial w_{pj}^T } &= \frac{\partial E_i}{\partial w_{pj}^T h} * \frac{\partial w_{pj}^T h }{\partial w_{pj}^T }
\end{align}

In the first part we substitute (\ref{eq:log}), rearrange the summation of derivatives, and apply the chain rule:
\begin{align}   \label{eq:dE-dwh}
    \frac{\partial E_i}{\partial w_{pj}^T h} 
    &= \frac{\partial}{\partial w_{pj}^T h} \Big ( - \sum_{q \in Q} \log P(m_q|q)  \Big ) \\
    &= - \sum_{q \in Q} \frac{\partial}{\partial w_{pj}^T h} \Big (  \log P(m_q|q)  \Big ) \\
    &= - \sum_{q \in Q} \frac{\partial \log P(m_q|q)}{\partial P(m_q|q)} *
    \frac{\partial P(m_q|q)}{\partial w_{pj}^T h}  \label{eq:dw_intermediate}
\end{align}

The derivative of the natural logarithm is trivial:
\begin{align} \label{eq:dlog}
    \frac{\partial \log P(m_q|q)}{\partial P(m_q|q)} &= \frac{1}{P(m_q|q)}
\end{align}

In the derivative of the probability of the correct child node belonging to a given parent node on the path to the correct class, with respect to the inner product $w_{pj}^T h$, we have to consider two things. First, if the parent node $p$ (to which the weight vector $w_{pj}$ belongs), is the same as the given parent node we consider $q$. Secondly, like in a regular softmax, whether the weight vector corresponds to the correct child node, i.e. if $j=m_q$:
\begin{align}
    \frac{\partial P(m_q|q)}{\partial w_{pj}^T h} &= 
    \begin{cases}
        P(m_q|q) ( \delta_{jm_q} -P(j|q)) & \quad \text{if } p=q \\
        0 & \quad \text{otherwise}
    \end{cases} \\
    &= \delta_{pq} P(m_q|q) ( \delta_{jm_q} -P(j|q)) \label{eq:dP_dwh}
\end{align}

The results of (\ref{eq:dlog})-(\ref{eq:dP_dwh}) are substituted in (\ref{eq:dw_intermediate}). The $P(m_q|q)$'s cancel out, and the -1 is brought inside the sum to rearrange $\delta_{jm_q} -P(j|q)$. In the $\sum_{q \in Q}$ we pass over all parent nodes that belong to the set $Q$ that make up the path to the correct class. Since we only consider one $p$, we check whether $p$ is in $Q$. The indicator function $\mathbbm{1}_{p \in Q}$ is one if $p$ is in $Q$ and zero otherwise. 
\begin{align}  
    \frac{\partial E_i}{\partial w_{pj}^T h} &= 
    - \sum_{q \in Q}  \frac{1}{P(m_q|q)} \delta_{pq} P(m_q|q) ( \delta_{jm_q} -P(j|q)) \\
    &= \sum_{q \in Q} \delta_{pq} (P(j|q)-\delta_{jm_q}) \\
    &=  \begin{cases}
            P(j|p)-\delta_{jm_p} & \quad \text{if } p \in Q \\
            0 & \quad \text{otherwise}
        \end{cases} \\
    &= \mathbbm{1}_{p \in Q} ( P(j|p)-\delta_{jm_p} ) \label{eq:dE_dwh}
\end{align}

The second part of (\ref{eq:chain_w}) is trivial:
\begin{align}   \label{eq:dwh-dw}
    \frac{\partial w_{pj}^T h}{\partial w_{pj}^T }   = h,
\end{align}

Combining the two results of (\ref{eq:dE_dwh}) and (\ref{eq:dwh-dw}) gives us:
\begin{align}
    \frac{\partial E_i}{\partial w_{pj}^T } 
    &= \mathbbm{1}_{p \in Q} (P(j|q)-\delta_{jm_p} ) h
\end{align}

This result means we only update weights that belong to the parent nodes that make up the path from the root to the correct class. Like in a regular softmax, the updates depend on whether we are updating the weight vector corresponding to the correct child node or an incorrect one.

\subsection{Derivative with respect to the hidden state}
The derivation of the Cross Entropy loss of observation $i$ with respect to the hidden state is to show that the network is updated using knowledge of the performance across the hierarchy of classes. 

\begin{align}
    \frac{\partial E_i}{\partial h } 
    &= \frac{\partial }{\partial h } \Big (  - \sum_{q \in Q} \log P(m_q|q) \Big ) \\
    &= - \sum_{q \in Q} \frac{\partial }{\partial h } \log P(m_q|q)\\
    &= - \sum_{q \in Q} \frac{\partial \log P(m_q|q)}{\partial P(m_q|q) } \frac{\partial  P(m_q|q)}{\partial h } \label{eq:h_1}
\end{align}

The first part of (\ref{eq:h_1}) is given in (\ref{eq:dlog}), the second part can calculated as follows:
\begin{align}
    \frac{\partial  P(m_q|q)}{\partial h }
    &= \sum_{j=1}^{J_q} \frac{\partial  P(m_q|q)}{\partial w_{qj}^T h } \frac{\partial w_{qj}^T h}{\partial h} \label{eq:sum_childs}
\end{align}

We have to consider all child nodes of parent $q$. Therefore we sum over all $J_q$ child nodes of parent $q$.  

\begin{align}
    \frac{\partial  P(m_q|q)}{\partial w_{qj}^T h } &= 
    P(m_q|q) (\delta_{jm_q} - P(j|q) ) \label{eq:triv1}
\end{align}

\begin{align} \label{eq:dwh-dh} 
    \frac{\partial w_{qj}^T h}{\partial h }   &= w_{qj}^T 
\end{align}
\\
Substituting the trivial derivatives of (\ref{eq:triv1})-(\ref{eq:dwh-dh}) into (\ref{eq:sum_childs}) gives us:
\begin{align}
    \frac{\partial  P(m_q|q)}{\partial h } &= 
    \sum_{j=1}^{J_q} P(m_q|q) (\delta_{jm_q} - P(j|q) ) w_{qj}^T \\
    &= P(m_q|q) \sum_{j=1}^{J_q} (\delta_{jm_q} - P(j|q) ) w_{qj}^T \label{eq:almost_final}
\end{align}
\\
Combining (\ref{eq:h_1}), (\ref{eq:dlog}), and (\ref{eq:almost_final}) results in:
\begin{align}
    \frac{\partial E_i}{\partial h } &=
    - \sum_{q \in Q} \frac{1}{ P(m_q|q)} P(m_q|q) \sum_{j=1}^{J_q} (\delta_{jm_q} - P(j|q) ) w_{qj}^T \\
    &= \sum_{q \in Q} \sum_{j=1}^{J_q} (\delta_{jm_q} - P(j|q) ) w_{qj}^T
\end{align}
\\
\indent The update of the hidden state (and therefore the rest of the network) is a combination of the performances across all child nodes that belong to the parent nodes that make up the path to the correct class. This shows that a neural network with a hierarchical softmax is truly a global hierarchical classifier.

\section{Hierarchical Structures} \label{sec:tax}
\begin{figure*}[htbp]
\includegraphics[width=0.8\textwidth]{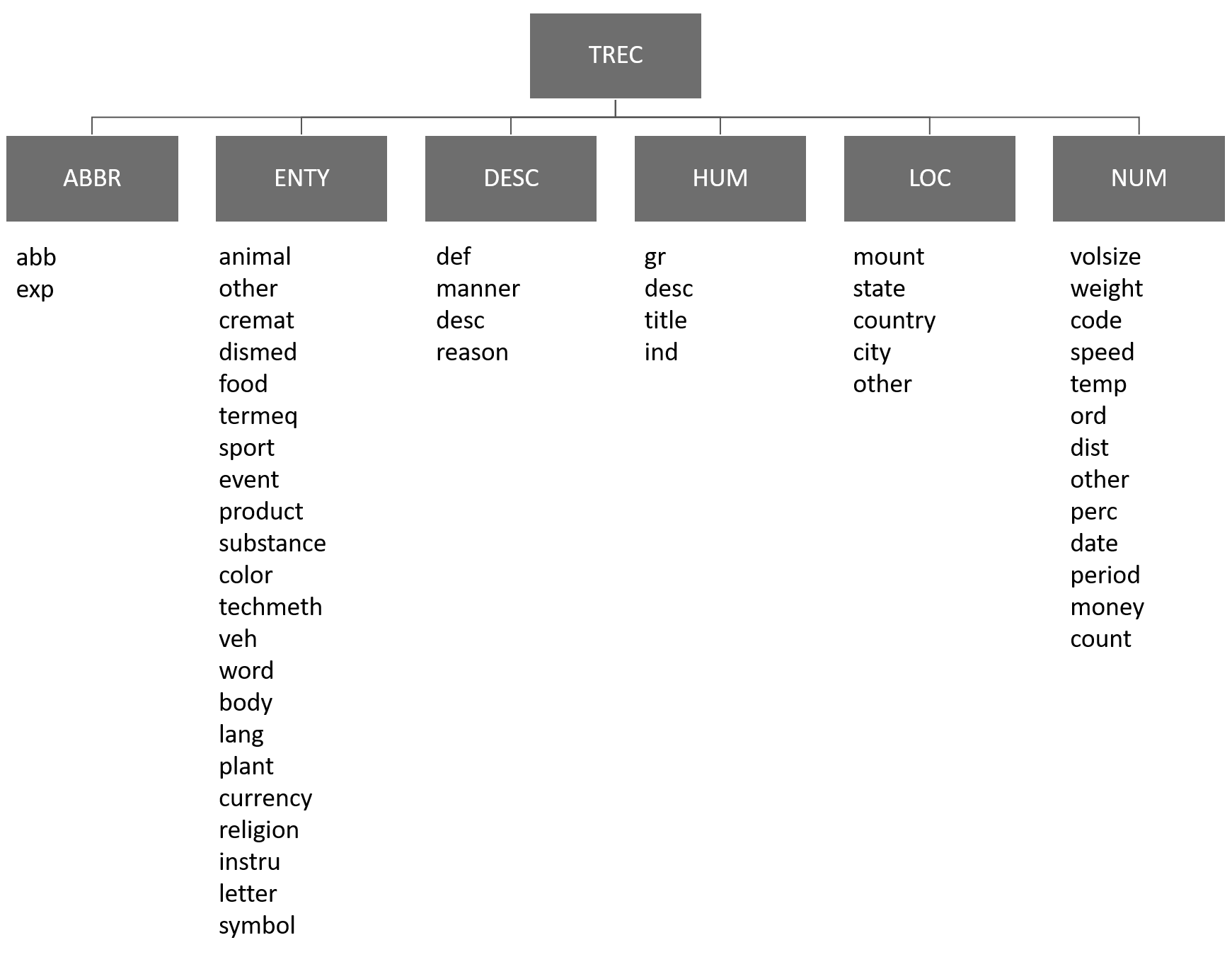}
\caption{The hierarchy introduced in the TREC dataset.}
\label{fig:trec}
\end{figure*}

\begin{figure*}[htbp]
\includegraphics[width=0.8\textwidth]{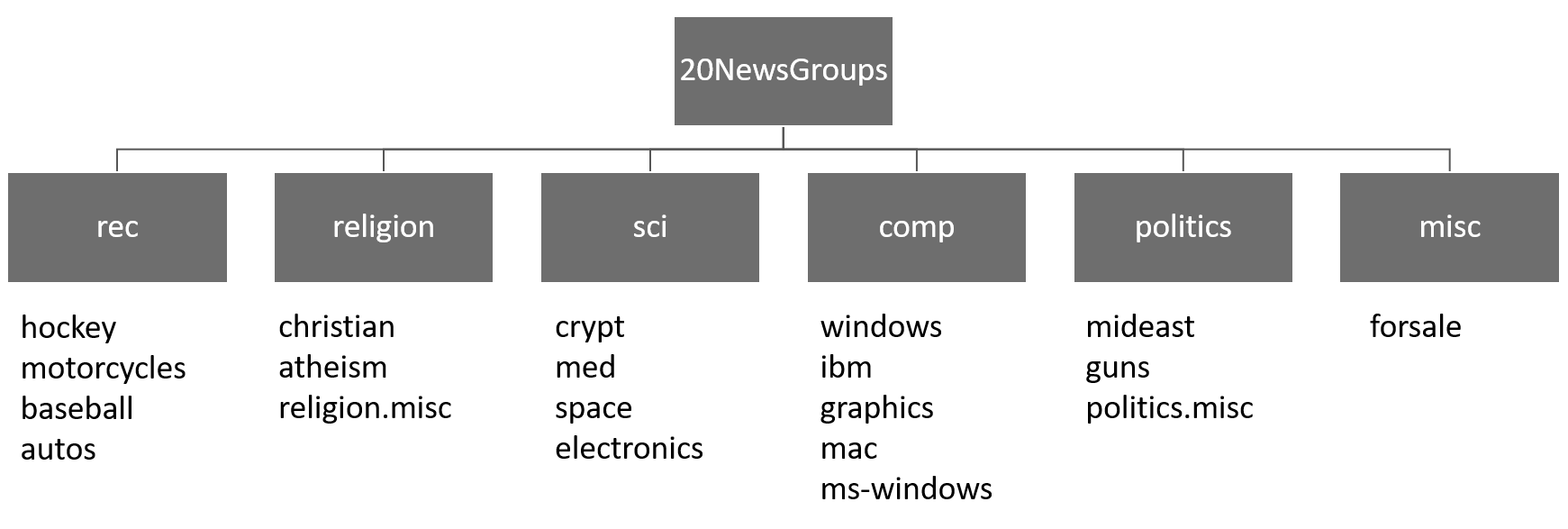}
\caption{The hierarchy introduced in the 20NewsGroups dataset.}
\label{fig:20ng}
\end{figure*}

\begin{figure*}[htbp]
\includegraphics[width=0.8\textwidth]{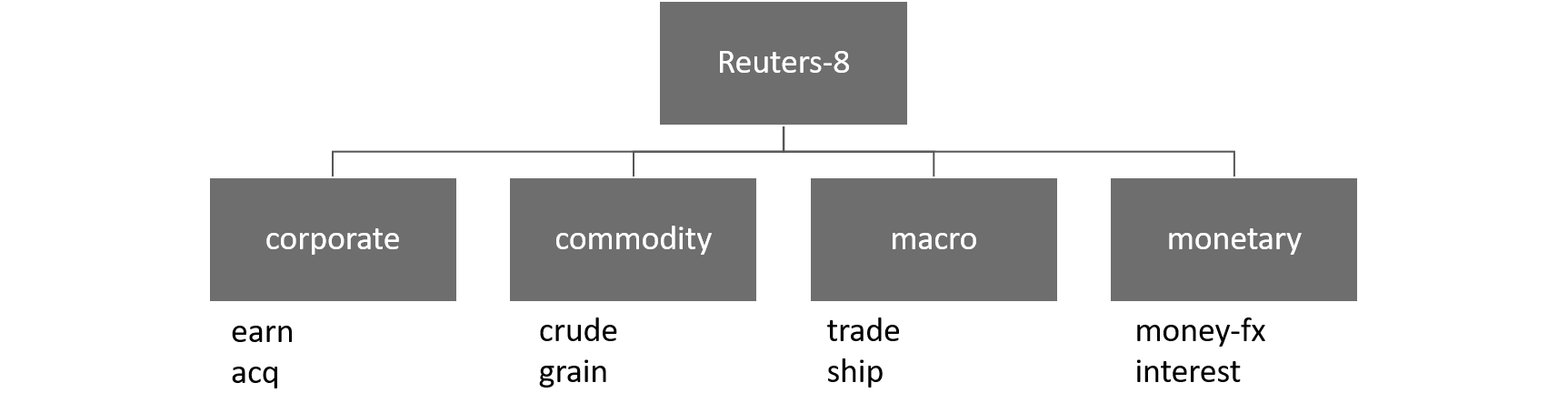}
\caption{The hierarchy introduced in the Reuters-8 dataset.}
\label{fig:r8}
\end{figure*}

\begin{figure*}[htbp]
\includegraphics[width=0.8\textwidth]{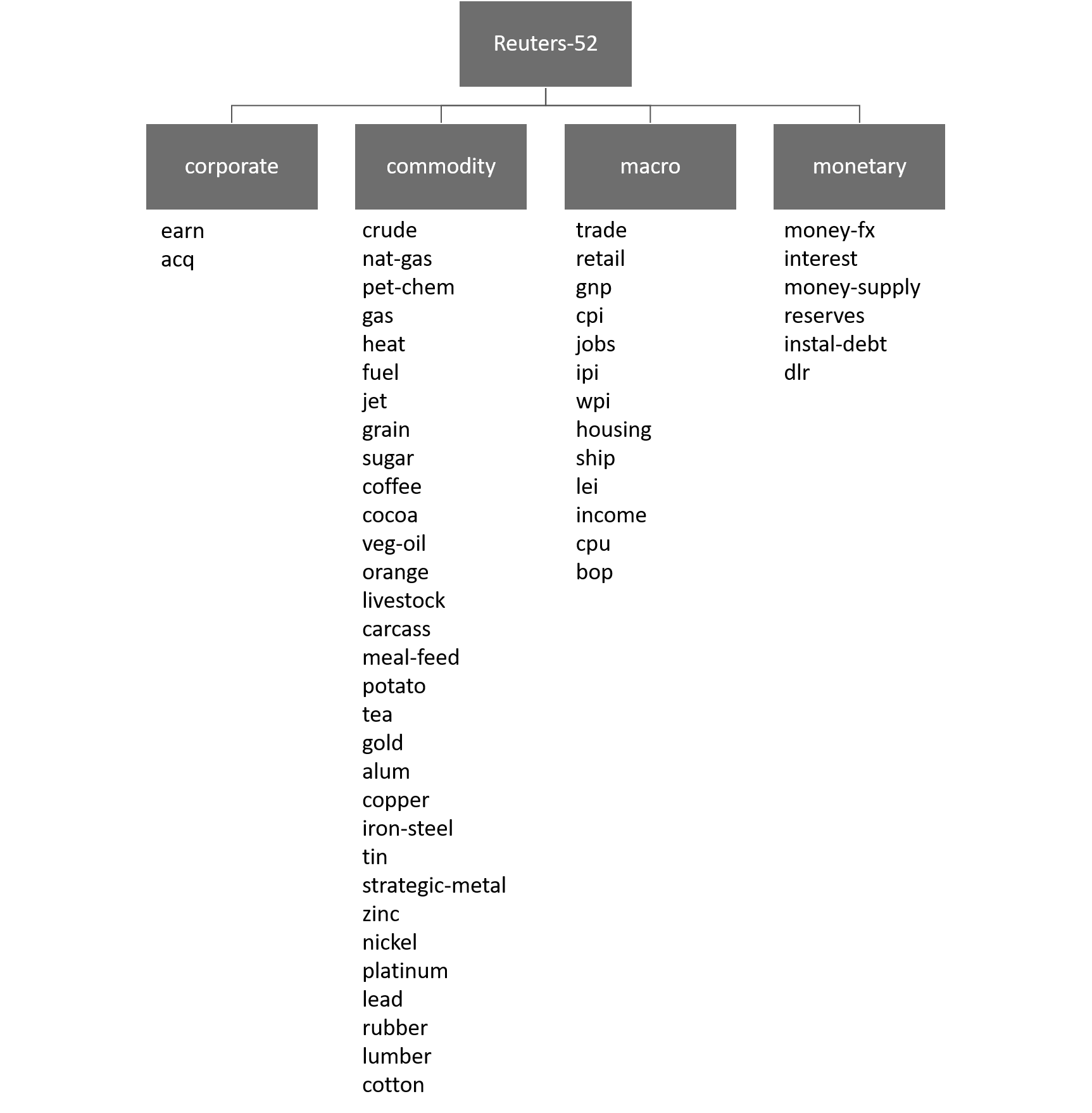}
\caption{The hierarchy introduced in the Reuters-52 dataset.}
\label{fig:r52}
\end{figure*}

This Appendix discloses the hierarchical structures used as class taxonomies. Figures \ref{fig:trec}-\ref{fig:r52} cover TREC, 20NewsGroups, Reuters-8, and Reuters-52 respectively. The name of the dataset represents the root node. This is connected to the categories we consider. Below the categories are the classes that belong to a respective category.

\end{document}